\documentclass[sigconf]{acmart}

\AtBeginDocument{%
  \providecommand\BibTeX{{%
    \normalfont B\kern-0.5em{\scshape i\kern-0.25em b}\kern-0.8em\TeX}}}

\usepackage{graphicx} 
\usepackage{float} 
\usepackage{subfigure} 
\usepackage{stfloats}
\usepackage{booktabs}
\usepackage{multirow}
\usepackage{bm}
\usepackage{color}
\usepackage{fancyhdr}
\usepackage{balance}
\copyrightyear{2020}
\acmYear{2020}
\setcopyright{acmcopyright}\acmConference[MM '20]{Proceedings of the 28th ACM International Conference on Multimedia}{October 12--16, 2020}{Seattle, WA, USA}
\acmBooktitle{Proceedings of the 28th ACM International Conference on Multimedia (MM '20), October 12--16, 2020, Seattle, WA, USA}
\acmPrice{15.00}
\acmDOI{10.1145/3394171.3414056}
\acmISBN{978-1-4503-7988-5/20/10}
\begin{document}
\fancyhead{}

\title{Black Re-ID: A Head-shoulder Descriptor for the Challenging Problem of Person Re-Identification}

\author{Boqiang Xu}
\authornote{This work is done when Boqiang Xu is an intern at JD AI Research.}
\affiliation{%
  \institution{University of Chinese Academy of Sciences}
}
\email{boqiang.xu@cripac.ia.ac.cn}

\author{Lingxiao He}
\authornote{Corresponding author.}
\affiliation{%
  \institution{ AI Research of JD.com}
}
\email{ helingxiao3@jd.com}

\author{Xingyu Liao}
\affiliation{%
  \institution{AI Research of JD.com}
}
\email{liaoxingyu5@jd.com}

\author{Wu Liu}
\authornote{Corresponding author.}
\affiliation{%
  \institution{AI Research of JD.com}
}
\email{liuwu1@jd.com}

\author{Zhenan Sun}
\affiliation{%
  \institution{Institute of Automation, Chinese of Academy of Sciences}
}
\email{znsun@nlpr.ia.ac.cn}

\author{Tao Mei}
\affiliation{%
  \institution{AI Research of JD.com}
}
\email{tmei@jd.com}

\renewcommand{\shortauthors}{Boqiang Xu and Lingxiao He, et al.}

\begin{abstract}
Person re-identification (Re-ID) aims at retrieving an input person image from a set of images captured by multiple cameras. Although recent Re-ID methods have made great success, most of them extract features in terms of the attributes of clothing (e.g., color, texture). However, it is common for people to wear black clothes or be captured by surveillance systems in low light illumination, in which cases the attributes of the clothing are severely missing. We call this problem the \emph{Black Re-ID} problem. To solve this problem, rather than relying on the clothing information, we propose to exploit head-shoulder features to assist person Re-ID. The head-shoulder adaptive attention network (HAA) is proposed to learn the head-shoulder feature and an innovative ensemble method is designed to enhance the generalization of our model. Given the input person image, the ensemble method would focus on the head-shoulder feature by assigning a larger weight if the individual insides the image is in black clothing. Due to the lack of a suitable benchmark dataset for studying the Black Re-ID problem, we also contribute the first Black-reID dataset, which contains 1274 identities in training set. Extensive evaluations on the Black-reID, Market1501 and DukeMTMC-reID datasets show that our model achieves the best result compared with the state-of-the-art Re-ID methods on both Black and conventional Re-ID problems. Furthermore, our method is also proved to be effective in dealing with person Re-ID in similar clothing. Our code and dataset are avaliable on \textcolor{
blue}{https://github.com/xbq1994/}.
\end{abstract}

\ccsdesc[500]{Computing methodologies~Visual content-based indexing and retrieval}
\ccsdesc[500]{Computing methodologies~Image representations}
\ccsdesc[500]{Computer systems organization~Neural networks}

\keywords{Black person re-identification, Head-shoulder descriptor, Adaptive attention}


\maketitle

\section{Introduction}
Person re-identification (Re-ID) is one of the main tasks in computer vision, with the purpose of retrieving the same person from overlapping cameras. In practice, it is common for people to wear black clothes or for clothes to appear black because captured by cameras in low light illumination, in which cases the attributes of the clothing will be lost considerably, posing great challenges to modern person re-identification algorithms.
\begin{figure}[t]
  \centering
  \includegraphics[width=\linewidth]{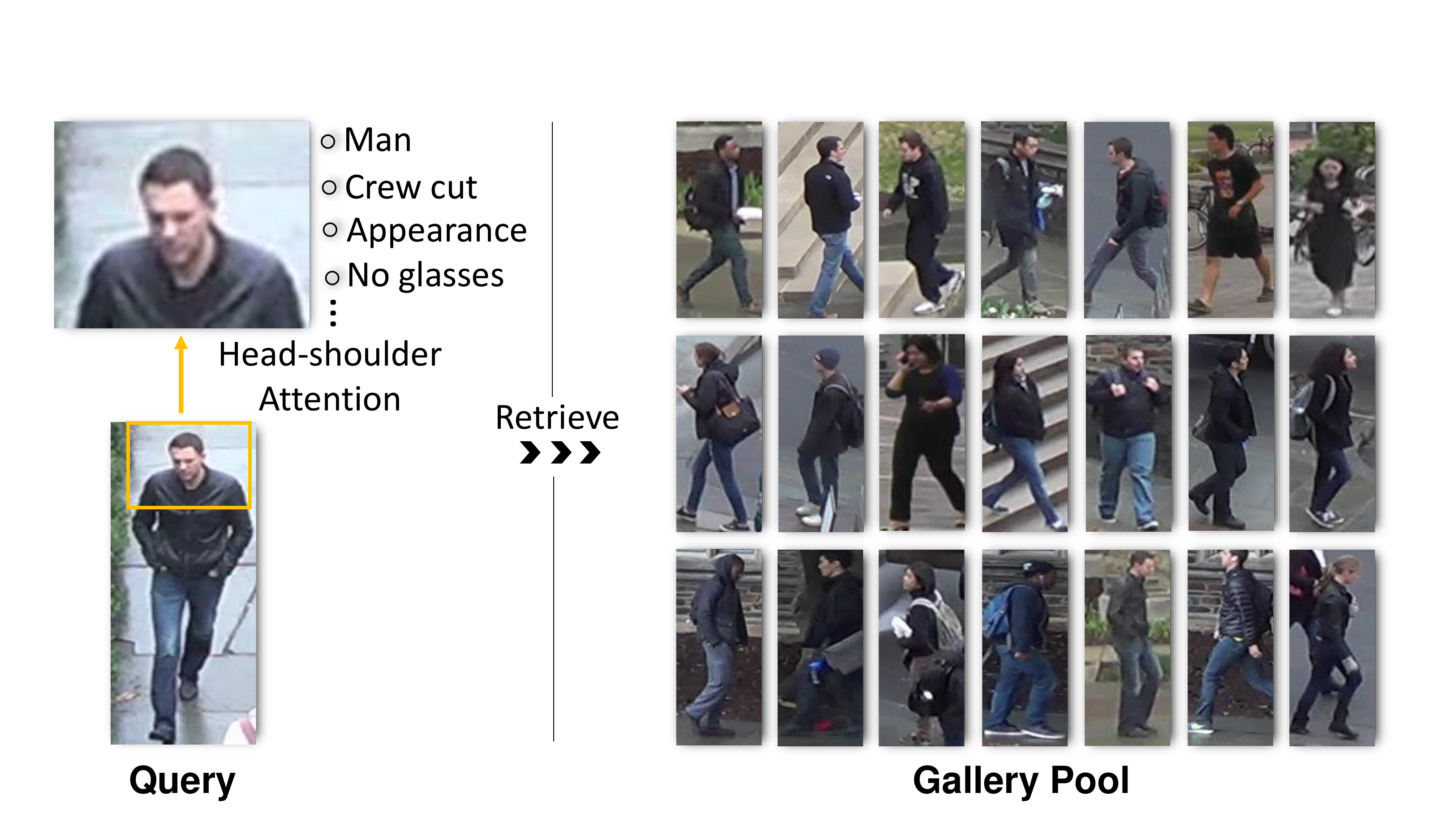}
  \caption{The illustration of the Black Re-ID problem. Black Re-ID poses a huge challenge to person re-identification due to the little information about clothes. Head-shoulder feature obtains abundant discriminative information for contributing the person Re-ID, such as gender, haircut, appearance and glasses.}
  \label{fig1}
\end{figure}
For example, as shown in Figure 1, when people are in black clothes or low light illumination, the seriously loss of clothing attributes causes great difficulties for person retrieval. It is very difficult to distinguish them based only on clothing appearance. We call the process of person Re-ID under black clothing or low light illumination the \emph{Black Re-ID} problem. In addition, We have counted the ids of the people in black 
clothing in the training set of Market1501 \cite{Zheng2015ScalablePR}, DukeMTMC-reID \cite{Ristani2016PerformanceMA} and CUHK03 \cite{DBLP:conf/cvpr/LiZXW14} in Table 1. From the result we can see that Black Re-ID problem is common in reality.

Recently, methods based on the deep learning networks have made great success in person re-identification by learning powerful person representations \cite{liu2019adaptive,Multi-Scale10.1145/2964284.2967209}. Deep learning based methods typically work on extracting a global feature of a person image  \cite{Zhang2016LearningAD}, which is robust but overlooking the intra-class variations (e.g. occlusions, poses, backgrounds). To alleviate this problem, the local features are leveraged to obtain a more discriminative representation for effective person Re-ID. Several pose-based methods \cite{Zhao2017SpindleNP,Zheng2019PoseInvariantEF} attend to extract local features from body parts such as arms, legs and torso. They utilize off-the-shelf pose estimators to enhance the attention to the corresponding local parts. Some part-based models \cite{Sun2018BeyondPM,Wang2018LearningDF} extract local detail information by slicing the feature into several horizontal grids, training them individually and aggregating them for a powerful representation. However, these methods extract features mainly rely on the attributes of the clothing (e.g. color, texture, style). In Black Re-ID problem, the information lost due to the black clothing or insufficient illumination hinders the performance of these Re-ID methods. As shown in Figure 1, it is very difficult to distinguish them based only on the clothes.

\begin{table}[]
\caption{The ids of the people in black clothing in the training set of Market1501 \cite{Zheng2015ScalablePR}, DukeMTMC-reID \cite{Ristani2016PerformanceMA} and CUHK03 \cite{DBLP:conf/cvpr/LiZXW14}}
\begin{tabular}{@{}cccc@{}}
\toprule
                     & Market1501 & DukeMTMC-reID & CUHK03 \\ \midrule
\# identities\_total & 751        & 702           & 767    \\
\# identities\_black & 98         & 347           & 331    \\ \bottomrule
\end{tabular}
\end{table}

In this paper, we propose to leverage the head-shoulder information to enhance the description of the feature for people in black clothing for an more effective person Re-ID. Head-shoulder part possesses abundant discriminative information, such as haircut, face and other appearances, for contributing the person Re-ID in solving the Black Re-ID problem. In particular, as shown in Figure 2, we design a two-stream network consists of the global stream and head-shoulder attention stream (HSA). For the global stream, it learns the global representation. For the head-shoulder attention stream (HSA), it learns the head-shoulder feature and utilizes the head-shoulder localization layer to take place of the off-the-shelf pose estimators with lightweight architecture. Furthermore, in order to make use of the head-shoulder information in solving both Black Re-ID and conventional problems, we propose the adaptive attention module to adapt the weights of the global and head-shoulder feature by the condition of the input person image. Specially, when the input person is in black clothing, our model would give more attention to the head-shoulder feature, compared to the person not in black clothes. This assembled method ensures that our model can make the most efficient use of the head-shoulder information in both the Black and conventional Re-ID conditions. In addition, we establish the first Black-reID dataset which contains 1274 identities in training set. Our experiment was conducted on the Black-reID dataset, Market1501 \cite{Zheng2015ScalablePR} and DukeMTMC-reID \cite{Ristani2016PerformanceMA}, demonstrating the advantage of our approach for person Re-ID on both Black and conventional  Re-ID problems.

The main contributions of this paper can be summarized as follows: 
\begin{itemize}
    
\item We firstly propose the study of the Black Re-ID problem and establish the first Black-reID dataset.
   
\item We propose the head-shoulder adaptive attention network (HAA), which makes use of the head-shoulder information to support person re-identification through the adaptive attention module. The HAA can be integrated with the most current Re-ID framework and is end-to-end trainable.
   
\item Our model is proved by the experiment that it is not only effective for Black Re-ID problem but also valid in similar clothing.
 
\item We achieve a new state of the art, and outperforms other Re-ID methods by a large margin in solving both Black Re-ID and conventional problems.
\end{itemize}

\section{Related Work}
\textbf{Person Re-ID.} The performance of person Re-ID has made great success recently \cite{he2019foreground-aware,10.1145/3343031.3350984,Li2016A,he2018deep}, due to the development of \textbf{CNN}s. Most of the \textbf{CNN}s based methods treat Re-ID as a classification task \cite{Zheng2017ADL}, aiming at divide person with same identity into the same category. To obtain a more discriminative and robust representation of the person, many methods \cite{Zhao2017SpindleNP,Zheng2019PoseInvariantEF,Zhao2017DeeplyLearnedPR,Sun2018BeyondPM,Wang2018LearningDF} integrate global feature with local features for an effective person re-identification. 

Local feature based methods enhance the discriminative capability of the final feature map. We briefly classify them into three categories: The first approach is pose-based Re-ID  \cite{Zhao2017SpindleNP,Zheng2019PoseInvariantEF}, which uses an off-the-shelf pose estimator to  extract
 the pose information for aligning body parts \cite{Suh2018PartAlignedBR} or generating person images \cite{Liu2018PoseTP}. However, the pose estimator is utilized through the whole training and testing process, making the network bigger and slower. The second approach is part-based Re-ID \cite{Sun2018BeyondPM,Wang2018LearningDF}, which slices image or global feature into several horizontal grids, training individually, and assembling for a discriminative person representation. However, this method is sensitive to the pose variations and occlusions, as they may compare corresponding part with different semantics. The third kind of methods leverage local information with attention maps \cite{Zhao2017DeeplyLearnedPR}, which can be trained under less supervisory signals compared to the pose annotations. It pays more attention to the regions of interest and robust to the background clutter, but nonetheless, the area of concern may not contain body parts. Our method belongs to the first category. In contrast to the other pose-based methods, we propose the $HSA$ stream to take place of the pose estimator, making the network lightweight. Furthermore, we introduce an adaptive attention module to decide the weight of global and local features by the condition of the input person image. 
 
 ~\\\noindent\textbf{Head-shoulder Information.} Head-shoulder information is reliable for retrieving as the corresponding region can usually be captured in reality, while the whole body may be occluded. Moreover, the head-shoulder feature possesses abundant discriminative information for contributing the person Re-ID, such as haircut, complexion or appearance. Unfortunately, as we know, there are few works \cite{Li2018MultiPoseLB} focusing on improving Re-ID performance with head-shoulder part. \citeauthor{Li2018MultiPoseLB} \cite{Li2018MultiPoseLB} propose a multi pose learning based model which takes head-shoulder part as input, aiming at solving partial Re-ID problem \cite{Zheng2015PartialPR} in crowded conditions. It tackles head-shoulder pose variations by learning an ensemble verification conditional probability distribution about relationship among multiple poses. However, it focuses mainly on pose variation, contributing little work to the head-shoulder localization, feature extraction and feature fusion with the global representation. Moreover, it performs worse than other methods (i.e., MGN \cite{Wang2018LearningDF}, PCB \cite{Sun2018BeyondPM}, AlignedReID \cite{Luo2019AlignedReIDDM}) when the whole body is available as it only takes head-shoulder part as input.

\begin{figure*}[t]
  \centering
  \includegraphics[width=\textwidth]{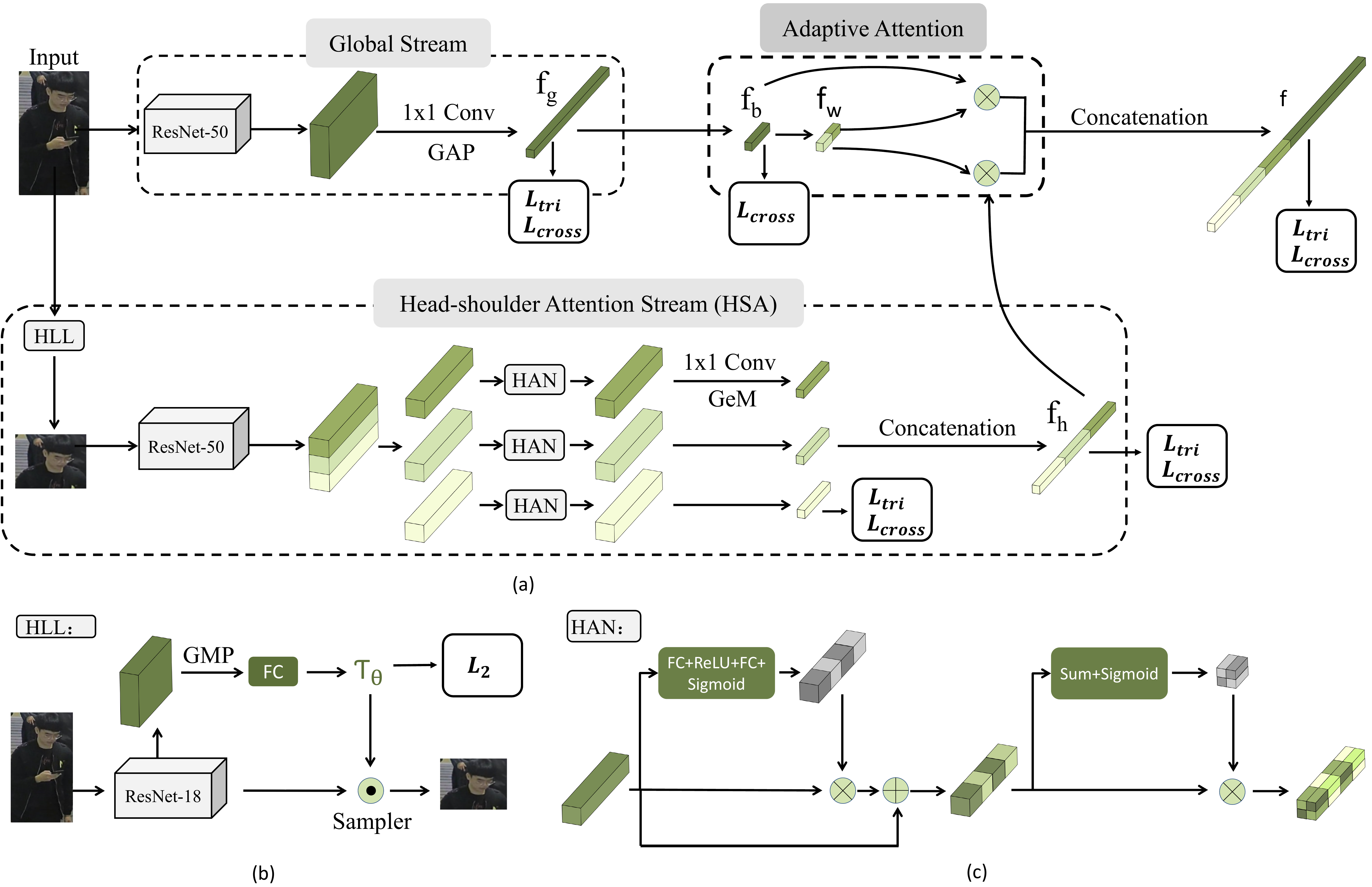}
  \caption{Overview of the proposed framework. (a) Our model consists of two streams: the global stream extracts the global feature from the input person image; the head-shoulder attention stream (HSA) crops the head-shoulder region by head-shoulder localization layer (HLL), which is divided into three horizontal stripes later and fed into head-shoulder attention network (HAN) to give the head-shoulder representation; at the end of the model, global and head-shoulder features are assembled through the adaptive attention module to produce the final representation for Re-ID. (b) The detailed structure of the head-shoulder localization layer (HLL). (c) The detailed structure of the head-shoulder attention network (HAN). Here, $\otimes$, $\oplus$, GAP, GMP, GeM, $L_{triplet}$, $L_
 {ce}$, $L_2$ indicate element-wise multiplication, element-wise addition, global average pooling, global max pooling, generalized mean pooling, triplet loss, cross-entropy loss and L2 loss respectively, and $\odot$ is a sampler.}
\end{figure*}

\section{Our Approach}
We show in Figure 2 (a) an overview of our framework. As our method could be integrated with the most current Re-ID model ($e.g.$, MGN \cite{Wang2018LearningDF}), the backbone can be selected according to different requirements of accuracy and speed. To illustrate our framework concisely, the ResNet-50 \cite{He2016DeepRL} trained for ImageNet classification \cite{Deng2009ImageNetAL} is exploited in demonstration. The network consists of two streams named global stream and head-shoulder attention stream (HSA). The first stream extracts the global feature from the person image. The second stream focuses on localizing and extracting head-shoulder information to make the final feature more discriminative for dealing with the Black Re-ID problem. Further more, we propose the adaptive attention module to adapt the weights of the global feature and head-shoulder feature, varying depending on whether a person is in black or not. We train our model end-to-end using cross-entropy, triplet and L2 losses. At test time, we extract features from person, and calculate the Euclidean distance between them to match people with the same ID. 

\subsection{Global Stream}
We extract global feature through a ResNet50 fasion network. Specifically, a feature map of size $C \times H \times W$ is extracted from a person image, where $C$, $H$, $W$ represent the number of channels, height and width, respectively. The resulting feature map is then processed by GAP and channel reduction to produce the global feature, denoted by $f_g$, of size $c \times 1 \times 1$, 

\subsection{Head-shoulder Attention Stream}
\textbf{Head-shoulder Feature Construction.} As mentioned in Section 2, several part-based Re-ID models attempt to localize body parts (e.g. legs, arms) with some off-the-shelf pose estimators \cite{Zhao2017SpindleNP,Zheng2019PoseInvariantEF} through both the training and testing phase. However, these methods make the Re-ID model bigger and slower, as it is a combination of two models. For the head-shoulder area, it's a less strictly defined area, and slight offset of localization would not affect Re-ID performance, which gives it the advantage of lower precision requirement of positioning than the segmentation task does. With these concerns, what we need is a lightweight localization layer that can learn a bounding box, representing the head-shoulder region. 

We propose the HSA stream to localize and extract the feature of the head-shoulder region, avoiding the use of pose estimators.  
HSA stream contains a head-shoulder localization layer (HLL), which is inspired by the success of $Spatial$ $Transformer$ $Network$ (\textbf{STN}) \cite{Jaderberg2015SpatialTN}. As illustrated in Figure 2 (b), HLL is an end-to-end trainable module with capacity of applying an affine transformation to the feature map, including scaling, translation and rotation. As the head-shoulder region is a simple bounding box, we only retain the ability to zoom and shift, equivalent to the cropping from the input. The transformation of the head-shoulder localization layer can be realized as follows:
$$\begin{pmatrix}x^s_i\\y^s_i\end{pmatrix}=\begin{bmatrix}s_x&0&t_x\\0&s_y&t_y\end{bmatrix}\begin{pmatrix}x^t_i\\y^t_i\\1\end{pmatrix}\eqno{(1)}$$
where $x^s_i$,$y^s_i$ and $x^t_i$,$y^t_i$ are the source positions and target coordinates of the $i$-th pixel respectively, $s_x$,$s_y$ are scaling parameters and $t_x$,$t_y$ are translation parameters. In the head-shoulder localization layer, the fully connected layer of size $C \times 4$ outputs $s_x$,$s_y$,$t_x$ and $t_y$, and then the following steps will sample pixels from the corresponding positions of the input person image to generate the bounding box.

Specifically, the input person images firstly pass through the head-shoulder localization layer, giving the head-shoulder region and resize to the same shape as input. Then, a feature map of size $C \times H \times W$ is extracted from the  head-shoulder region and sliced into 3 horizontal grids. The head-shoulder attention network(HAN) is applied to each individual horizontal slice, which are concatenated finally to produce the head-shoulder feature $f_h$ of size $c \times 1 \times 1$ .

~\\\noindent\textbf{Head-shoulder Attention Network.}
Figure 2 (c) shows the detailed structure of the HAN. Since different channels of feature maps represent different meanings, that is, the contributions of features to recognition vary from channel to channel, and different spatial location of features has diverse semantics. We introduce an attention network to enhance the representation of head-shoulder in both channel and spatial dimensions. 

Specifically, for the $i$-th($i=1,2,3$) slice, the input feature \textbf{$X_i$} passes through a gating mechanism, including a generalized mean pooling (GeM) \cite{Gu2018AttentionawareGM}, a fully connected layer with weight \textbf{$W_i$}$ \in{\mathbb{R}^{C\times{\frac{C}{r}}}}$ for dimension reduction, a ReLU activation, another fully connected layer with weight \textbf{$U_i$}$ \in{\mathbb{R}^{\frac{C}{r}\times{C}}}$ for dimension incrementation and a sigmoid activation $\sigma$. Here, $r$ is the reduction ratio. Then, the channels are reweighted by a shortcut connection architecture with  element-wise addition, which can be formulated as:
$$A_i=X_i+X_i \bullet d_i\eqno{(2)}$$
$$d_i=\sigma(U_iRELU(W_iX_i))\eqno{(3)}$$
where $|\bullet|$ is element-wise multiplication and \textbf{$A_i$} is the output after channel attention. For the description, \textbf{$A_i$} can be written as:
$$[A_{i}^1,A_{i}^2,A_{i}^3,...A_{i}^C]\eqno{(4)}$$
where \textbf{$A_{i}^n$}($n=1,2,...C$) are the features of each channel of \textbf{$A_i$}. The spatial attention is conducted by strengthening the peak responses, this process can be formulated as:
$$f_{hi}=A_i \bullet {\xi({\sum_{n=1}^{C}A_{i}^n})}\eqno{(5)}$$

\subsection{Adaptive Attention}
Most of the existing Re-ID methods \cite{Sun2018BeyondPM,Wang2018LearningDF}  directly concatenate global and local features, ignoring the relationship between the weight of the features and input conditions. That is, no matter what kind of person is input such as occluded or exposed, the network gives the same attention to the global feature as to the local feature. To alleviate this problem, we propose the adaptive attention module to determinate the global and local feature weights by distinguishing input types. Specifically, the adaptive attention stream would decide if it is a person in black first, and give more attention to head-shoulder feats for people in black clothes than that not in black.

Concretely, firstly, the global feature $f_g$ is fed into a fully connected layer to gather the feature of size $N \times 2$, where $N$ is the batch size, denoted by $f_b$, representing whether the input person is in black. After that, $f_b$ is fed into another fully connected layer, giving feature map $f_w$  of size $N \times 2$. $f_w$ is the weights of global feature and head-shoulder feature, varying depending on whether the person is in black or not. That is, higher attention would be applied to the head-shoulder feature when the person is in black clothes. Finally, we integrate the global feature and head-shoulder feature as follows:
$$f = (f_g \bullet w_1) \circledast (f_h \bullet w_2)\eqno{(6)}$$
where \textbf{$f_w$}$=\begin{bmatrix}w_1&w_2\end{bmatrix}$, $|\bullet|$ is element-wise multiplication, $f_g$ and $f_h$ are global feature and head-shoulder feature respectively and $\circledast$ means concatenate method. The feature $f$ is used as the person representation for Re-ID.

\subsection{Model Training}
To train our model, we use triplet and cross-entropy loss, balanced by the parameter $\alpha$ and $\beta$ as follows:
$$\mathcal{L} = \alpha \mathcal{L}_{triplet} + \beta \mathcal{L}_{ce}$$
where we denote by $\mathcal{L}_{triplet}$ and $\mathcal{L}_{ce}$, triplet and cross-entropy losses, respectively. The cross-entropy loss is defined as 
$$\mathcal{L}_{ce}=-\sum_{i=1}^{N}\log\frac{\exp(\textbf{W}_{y_i}\textbf{h}_p^i+b_{y_i})}{\sum_{k=1}^C\exp(\textbf{W}_k\textbf{h}_p^i+b_k)}\eqno{(8)}$$
where $N$ is the number of images in mini-batch, $y_i$ is the label of feature $h_p^i$ and $C$ is the number of classes. Given a triplet consisting of the anchor, positive, and negative features (i.e., $\textbf{q}_{i,j}^A$, $\textbf{q}_{i,j}^P$ and $\textbf{q}_{i,j}^N$), the batch-hard triplet loss \cite{Hermans2017InDO} is formulated as follows:
$$\mathcal{L}_{triplet}=\sum_{k=1}^{N_k}\sum_{m=1}^{N_M}[\alpha + \mathop{\max}_{n=1...M}||\textbf{q}_{k,m}^A-\textbf{q}_{k,n}^P||_2-\mathop{\min}_{\substack{l=1...K\\n=1...N\\l\ne k}}||\textbf{q}_{k,m}^A-\textbf{q}_{l,n}^N||_2]_+\eqno{(9)}$$
where $\alpha$ denotes the margin. To supervise the study of the head-shoulder localization layer and minimize the regression error of it, L2-loss is adopted and defined as follows:
$$\mathcal{L}_2=\frac{1}{2N}\sum_{i}||l^{(i)}-r(h^{i})||_2^2\eqno{(10)}$$
where $N$ is the batch size, $l^{(i)}$ and $h^{(i)}$ are the ground-truth label and the prediction of the $i$-th bounding box of the head-shoulder region and $r$ is the function that transform $h^{(i)}$ to the same coordinate system as $l^{(i)}$.

\section{The Black-reID Dataset}
To promote the research on the Black Re-ID problem, we introduce Black-reID, the first dataset focusing on Black Re-ID problem, derived from the Market1501 \cite{Zheng2015ScalablePR}, DukeMTMC-reID \cite{Ristani2016PerformanceMA}, Partial \cite{7410888} and Occluded \cite{Zhuo2018OccludedPR} datasets. 

\subsection{Properties of Black-reID Dataset}
Some examples of the Black-reID dataset are shown in Figure 3. There are a few advantages to make Black-reID appealing. First, it is the first dataset for Black Re-ID problem. According to the clothes that the people wear, the Black-reID dataset consists of two groups. The first group is black group, which focuses on the Black-reID problem and contains 5,649 images covering 688 identities in training set and 1,644 identities of 1,489 query images and 4,973 gallery images in test set. The second group is white group, which is constructed for investigating the Re-ID ability of the model in similar clothing. The white group contains 10,040 images of 586 subjects for training, 2,756 images of 628 subjects for query and 10336 images for gallery. Second, to fit the reality, both two groups include people wearing the corresponding color and those who are not. The reason for this design is that we want our method can not only solve the Black Re-ID problem effectively, but also be reliable when dealing with conventional scenarios. Third, we have labeled the bounding box of head-shoulder region for the training set in Black-reID and annotated ids of person in black and white clothes for the two groups, respectively.

\subsection{Data Collection}
We pick out 98 ids, 347 ids, 5 ids and 42 ids who are in black clothes from Market1501 \cite{Zheng2015ScalablePR}, DukeMTMC-reID \cite{Ristani2016PerformanceMA}, Partial \cite{7410888} and Occluded \cite{Zhuo2018OccludedPR} datasets respectively for the black group, and 665 pedestrians who are in white from Market1501 \cite{Zheng2015ScalablePR} for the white group. Moreover, we randomly add some people in other clothes to these two groups for creating the Black-reID dataset. 

\begin{figure}[h]
  \centering
  \includegraphics[width=\linewidth]{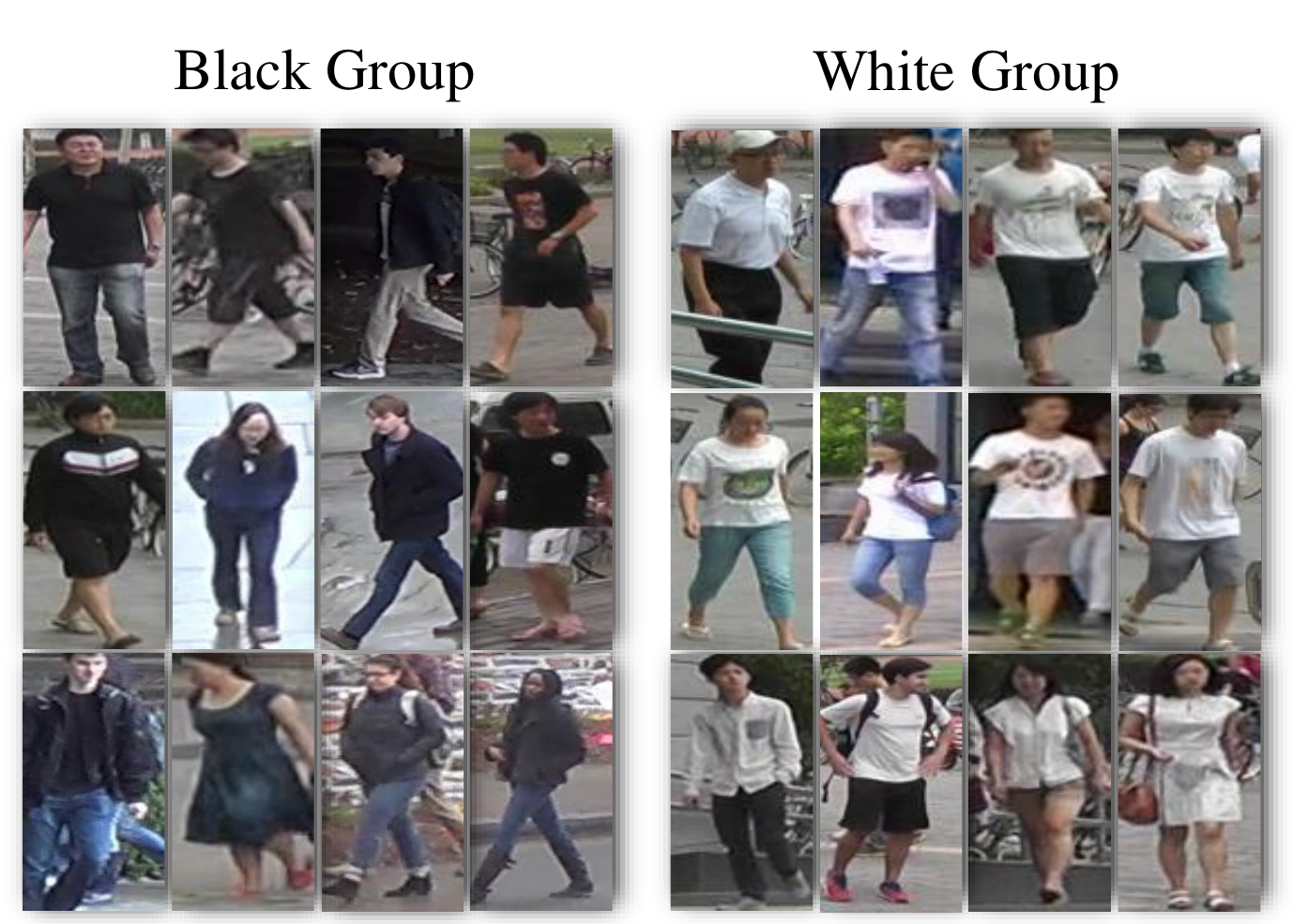}
  \caption{Some examples of the Black-reID dataset}
\end{figure}

\section{Experiment}
\subsection{Implementation details}
\textbf{Dataset} Our method is tested on the following three datasets and compare with the state-of-the-art methods. 1) The Black-reID dataset established by us, which contains 1274 identities in training set. 2) The Market1501 dataset \cite{Zheng2015ScalablePR} consists of 32,688 images of 1,501 subjects captured by six cameras. 3) The DukeMTMC-reID \cite{Ristani2016PerformanceMA} offers 16,522 training images of 702 identities, 2,228 query and 17,661 gallery images of 702 identities. Additionally, we manually labeled the head-shoulder bounding box for the Market1501 and DukeMTMC-reID datasets.

~\\\noindent\textbf{Training.} We resize all the training images into 384 $\times$ 128. We set the number of feature channels $c$ to 1536 and batch size $N$ to 64. We adopt horizontal flipping and random erasing \cite{Zhong2017RandomED} for data augmentation. Following the work of \cite{Sun2018BeyondPM}, the GAP and fully connected layers are removed from the original ResNet-50 architecture and the stride of the last convolutional layer is set to 1. Firstly, we pretrain the head-shoulder localization layer on  Black-reID dataset and freeze it in the following training. Next, we train the global stream and head-shoulder attention stream for 50 epochs individually, and then train them together with adaptive attention module for another 40 epochs. We use the adaptive gradient (\textbf{Adam}) \cite{Kingma2015AdamAM} as the optimizer with $\beta_1$ of 0.9, $\beta_1$ of 0.999 and weight decay of 5e-4. The learning rate is initially set to 3e-4 and divided by 10 at 40 and 70 epochs. We set the weight parameter $\alpha$ and $\beta$ to 1,1 and 1,2 for the integration of our model with ResNet50 \cite{He2016DeepRL} and MGN \cite{Wang2018LearningDF}, respectively.

~\\\noindent\textbf{Evaluation Metrics.} We utilize mean average precision (mAP) and Cumulative Matching Characteristic (CMC) curves to evaluate the performance of various Re-ID models. All the experiments are conducted in a single query setting.

\begin{table}[htbp]
\small
\center
\setlength{\tabcolsep}{4mm}
\caption{Quantitative comparison with the state-of-the-art methods in person re-id on Black-reID dataset. Bold number denote the best performance. We denote HAA (ResNet50) and HAA (MGN) by the method selecting ResNet50 and MGN as the backbone respectively.}
\begin{tabular}{@{}lcccc@{}}
\toprule
\multirow{2}{*}{Method} & \multicolumn{2}{c}{Black Group} & \multicolumn{2}{c}{White Group} \\
                        & mAP            & Rank-1         & mAP            & Rank-1         \\ \midrule
ResNet50 \cite{He2016DeepRL}                & 70.8           & 80.9           & 75.8           & 89.5           \\
PCB \cite{Sun2018BeyondPM}                    & 73.4           & 83.2           & 78.2           & 90.8           \\
AlignedReID \cite{Zhang2017AlignedReIDSH}            & 75.5           & 83.5           & 80.5           & 91.3           \\
MGN \cite{Wang2018LearningDF}                    & 79.1           & 86.7           & 85.8           & 94.3           \\ \midrule
HAA (ResNet50)           & 79.0           & 86.7           & 84.4           & 93.5           \\
HAA (MGN)                & \textbf{83.8}  & \textbf{91.0}  & \textbf{88.1}  & \textbf{95.3}  \\ \bottomrule
\end{tabular}
\end{table}

\subsection{Comparison with the state of the art}
\noindent\textbf{Results on Black-reID.} 
We compare in Table 2 our method with the state-of-the-art Re-ID models (i.e., MGN \cite{Wang2018LearningDF}, PCB \cite{Sun2018BeyondPM},\linebreak AlignedReID \cite{Luo2019AlignedReIDDM}) on Black-reID dataset. We denote HAA (ResNet50) and HAA (MGN) by the method selecting ResNet50 and MGN as the backbone respectively. Table 2 shows that HAA (ResNet50) and HAA (MGN) outperform their control groups (i.e., ResNet50 and MGN, respectively) by a large margin. In the black group, HAA (MGN) achieves the best result with mAP of $83.8\%$ and rank-1 accuracy of $91\%$, which is $4.7\%$ and $4.3\%$ higher than the corresponding metrics of MGN respectively. HAA (ResNet50) also gives $8.2\%$ and $5.8\%$ higher points in mAP and rank-1 than ResNet50, which reaches the same level as MGN. This demonstrates the effectiveness of our HAA in dealing with the Black Re-ID problem. In the white group, we can see that HAA (ResNet50) gives $8.6\%$ and $4\%$  higher points in mAP and rank-1 accuracy than ResNet50 while HAA (MGN) achieves the best result with mAP of $88.1\%$ and rank-1 accuracy of $95.3\%$, which is $2.3\%$ and $1\%$ higher than the corresponding metrics of MGN respectively. The result proves that our model is not only effective for Black Re-ID problem but also valid in solving similar clothes. We compare some retrieved results between PCB \cite{Sun2018BeyondPM}, AlignedReID \cite{Luo2019AlignedReIDDM}, MGN \cite{Wang2018LearningDF} and our method in Figure 4. As shown in Figure 4, PCB \cite{Sun2018BeyondPM}, AlignedReID \cite{Luo2019AlignedReIDDM} and MGN \cite{Wang2018LearningDF} extracts features mainly rely on the clothing attributes. By contrast, we can see that our proposed HAA can improve the Re-ID performance on Black Re-ID problem by taking advantage of the head-shoulder attributes to make the representation more discriminative. Furthermore, the Re-ID performance in Black -reID dataset is lower than that in Market1501 and DukeMTMC-reID. This proves that when people are in black clothes, lacking of attributes of clothing, the Re-ID performance of the model is really degraded.

\begin{table*}[t]
\center
\setlength{\tabcolsep}{9mm}
\caption{Performance (\%) comparisons with the state-of-the-art methods on Market1501 and DukeMTMC-reID.}
\begin{tabular}{@{}llcccc@{}}
\toprule
\multirow{2}{*}{}                   & \multirow{2}{*}{Method} & \multicolumn{2}{c}{Market1501} & \multicolumn{2}{c}{DukeMTMC-reID} \\
                                    &                         & mAP           & Rank-1         & mAP             & Rank-1          \\ \midrule
Basic-CNN                           & ResNet50 \cite{He2016DeepRL}                & 84.6          & 93.3           & 75.3            & 86.2            \\ \midrule
\multirow{12}{*}{Pose-guided methods} & Spindle \cite{Zhao2017SpindleNP}                 & -             & 76.9           & -               & -               \\
                                    & PIE \cite{Zheng2019PoseInvariantEF}                     & 53.9          & 78.7           & -               & -               \\
                                    & MSCAN \cite{Li2017LearningDC}                   & 57.5          & 80.8           & -               & -               \\
                                    & PDC \cite{Su2017PoseDrivenDC}                     & 63.4          & 84.1           & -               & -               \\
                                    & Pose Transfer \cite{Liu2018PoseTP}           & 68.9          & 87.7           & 48.1            & 68.6            \\
                                    & PN-GAN \cite{Qian2018PoseNormalizedIG}                  & 72.6          & 89.4           & 53.2            & 73.6            \\
                                    & PSE \cite{Sarfraz2018APE}                     & 69.0            & 87.7           & 62.0              & 79.8            \\
                                    & MGCAM \cite{Song2018MaskGuidedCA}                   & 74.3          & 83.8           & -               & -               \\
                                    & MaskReID \cite{Qi2018MaskReIDAM}                & 75.3          & 90.0           & 61.9            & 78.9            \\
                                    & Part-Aligned \cite{Suh2018PartAlignedBR}            & 79.6          & 91.7           & 84.4            & 69.3            \\
                                    & AACN \cite{Xu2018AttentionAwareCN}                    & 66.9          & 85.9           & 59.3            & 76.8            \\
                                    & SPReID \cite{Kalayeh2018HumanSP}                  & 81.3          & 92.5           & 71.0            & 84.4            \\ \midrule
\multirow{4}{*}{Part-based methods}       & AlignedReID \cite{Zhang2017AlignedReIDSH}             & 79.3          & 91.8           & -               & -               \\
                                    & Deep-Person \cite{Bai2020DeepPersonLD}             & 79.6          & 92.3           & 64.8            & 80.9            \\
                                    & PCB+RPP \cite{Sun2018BeyondPM}                 & 81.6          & 93.8           & 69.2            & 83.3            \\
                                    & MGN \cite{Wang2018LearningDF}                    & 86.9          & 95.7           & 78.4            & 88.7            \\ \midrule
\multirow{4}{*}{Attention-based methods}    & DLPAP \cite{Zhao2017DeeplyLearnedPR}                   & 63.4          & 81.0           & -               & -               \\
                                    & HA-CNN \cite{Li2018HarmoniousAN}                  & 75.7          & 91.2           & 63.8            & 80.5            \\
                                    & DuATM \cite{Si2018DualAM}                   & 76.6          & 91.4           & 64.6            & 81.8            \\
                                    & Mancs \cite{Wang2018MancsAM}                   & 82.3          & 93.1           & 71.8            & 84.9            \\ \midrule
                                    & HAA (Ours)                & \textbf{89.5}   & \textbf{95.8}  & \textbf{80.4}   & \textbf{89.0}            \\ \bottomrule
\end{tabular}
\end{table*}

~\\\noindent\textbf{Results on Market1501 and DukeMTMC-reID.} We also compare our proposed method with current state-of-the-art methods of three categories on Market1501 and DukeMTMC-reID in Table 3. $Pose-guided$ $methods$ leverage pose information to extract more discriminative local details or align features. $Part-based$ $methods$ slice the image/feature maps into several horizontal grids to assist Re-ID. $Attention-based$ $methods$ compute attention maps to consider discriminative regions of interest. Note that we do not utilize re-ranking \cite{Zhong2017RerankingPR} in all our results for clear comparisons. 

From Table 3, we can see that HAA (Ours) gives the best results on all datasets and specifically gives big improvement to mAP. HAA (Ours) achieves mAP of $89.5\%$ and rank-1 accuracy of $95.8\%$ on the Market1501, which is $2.6\%$ and $0.1\%$ higher than the corresponding metrics of MGN respectively, and mAP of $80.4\%$ and rank-1 accuracy of $89\%$ for the DukeMTMC-reID, which is $2\%$ and $0.3\%$ higher than the MGN respectively.

\begin{table*}[t]
\center
\setlength{\tabcolsep}{6.5mm}
\caption{ Ablation study of the adaptive attention module on Black-reID, Market1501 and DukeMTMC-reID datasets. w/o: without.}
\begin{tabular}{@{}lcccclc@{}}
\toprule
\multirow{2}{*}{Methods} & \multicolumn{2}{c}{Black-reID} & \multicolumn{2}{c}{Market1501} & \multicolumn{2}{c}{DukeMTMC-reID} \\ \cmidrule(l){2-7} 
                         & mAP           & Rank-1         & mAP           & Rank-1         & mAP            & Rank-1           \\ \midrule
HAA (ResNet50) (w/o adaptive attention)        & 78.3          & 86.2           & 85.3          & 93.8           & 73.1           & 85.6             \\
HAA (ResNet50) (adaptive attention)        & \textbf{79.0}            & \textbf{86.7}           & \textbf{85.6}          & \textbf{94.2}           & \textbf{74.2}           & \textbf{86.4}             \\ \bottomrule
\end{tabular}
\end{table*}

\begin{table}[t]
\center
\setlength{\tabcolsep}{7mm}
\caption{Ablation study of the pooling methods on Black-reID dataset. GAP, GMP, GeM indicate global average pooling, global max pooling and generalized mean pooling, respectively.}
\begin{tabular}{@{}lcc@{}}
\toprule
                           & mAP         & Rank-1        \\ \midrule
ResNet50 \cite{He2016DeepRL}                   & 70.8        & 80.9          \\
HAA (ResNet50) (GAP)         & 75.2        & 82.7          \\
HAA (ResNet50) (GMP)         & 77.3        & 85.3          \\
HAA (ResNet50) (GeM Pooling) & \textbf{79.0} & \textbf{86.7} \\ \bottomrule
\end{tabular}
\end{table}

\begin{table}[t]
\center
\setlength{\tabcolsep}{4.5mm}
\caption{Quantitative comparison of diverse features on
the Black-reID dataset}
\begin{tabular}{@{}lcc@{}}
\toprule
                     & mAP         & Rank-1        \\ \midrule
ResNet50 \cite{He2016DeepRL}              & 70.8        & 80.9          \\    
HAA (ResNet50) (Global)               & 78.6        & 86.4          \\
HAA (ResNet50) (Head-shoulder)        & 44.8        & 51.5          \\
HAA (ResNet50) (Global+Head-shoulder) & \textbf{79.0} & \textbf{86.7} \\ \bottomrule
\end{tabular}
\end{table}

\begin{figure*}[t]
  \centering
  \includegraphics[width=\linewidth,height=12cm]{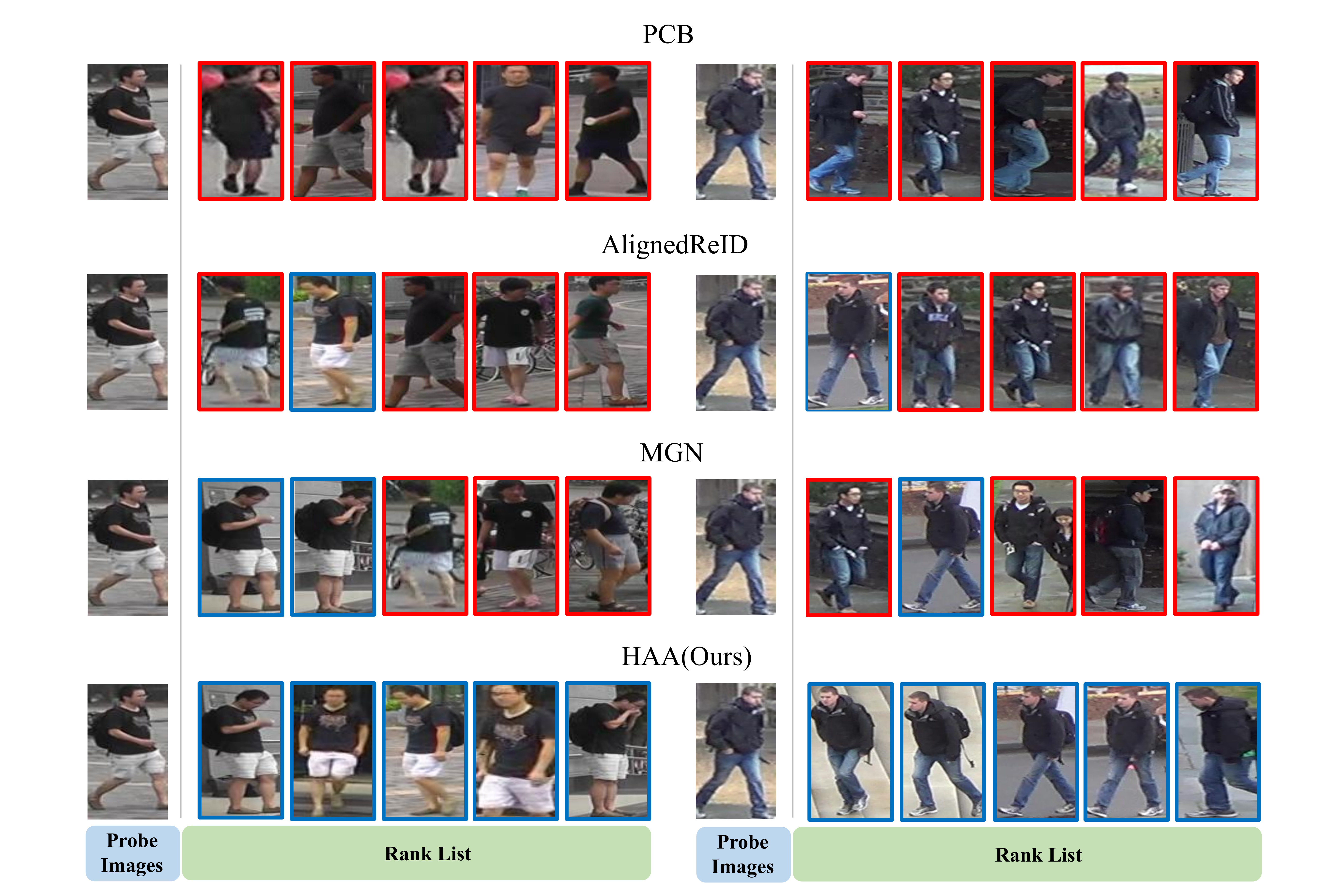}
  \caption{Comparison of the PCB \cite{Sun2018BeyondPM}, AlignedReID \cite{Luo2019AlignedReIDDM}, MGN \cite{Wang2018LearningDF} and our proposed HAA in solving Black Re-ID problem. Blue and red rectangles indicate correct and error retrieval results,respectively. More blue/correct retrievals in HAA result show that the head-shoulder information is discriminative for better Re-ID results.}
\end{figure*}

\section{Ablation Study}
We conduct extensive ablation studies on the black group of the Black-reID dataset based on the HAA (ResNet50).
  
~\\\noindent\textbf{The Impact of the Adaptive Attention Module.} In our method, we concatenate the global and head-shoulder feature through the adaptive attention module. To show how the ensemble method works, we set up a control group where the global and head-shoulder feature are directly concatenate during the whole training and testing process. The result is illustrated in Table 4. We denote HAA (ResNet50) (w/o adaptive attention) and HAA (ResNet50) (adaptive attention) by directly concatenate the global and head-shoulder feature and concatenate them through adaptive attention module, respectively. From the results, we can see that the adaptive attention stream gives the performance gains of 0.7$\%$ and 0.5$\%$ for mAP and rank-1 accuracy on Black-reID, 0.3$\%$ and 0.4$\%$ for mAP and rank-1 accuracy on Market1501 and 1.1$\%$ and 0.8$\%$ for mAP and rank-1 accuracy on DukeMTMC-reID,respectively. The result proves the effectiveness of the adaptive attention module in solving both Black and conventional Re-ID problems.

~\\\noindent\textbf{Ablation Study of the GeM Pooling.} We adopt GeM pooling in the HSA stream. To compare the effectiveness of diverse pooling methods, we replace the GeM pooling with GAP and GMP, and conduct the experiments on Black-reID dataset in Table 5. From the third and fourth rows, we can see that GMP gives better results than GAP, the reason is that GAP covers the whole person image and is easily distracted by background and black-clad while GMP gets over this problem by aggregating the feature from the most discriminative part. The results in the next row show that GeM pooling achieve the best result, giving the performance gains of 1.7$\%$, 1.4$\%$ and 3.8$\%$, 4$\%$ for mAP and rank-1 accuracy than GMP and GAP respectively. GeM pooling is given by:
$$e=[(\sum_{u \in \sigma}x_{cu}^p)^{\frac{1}{p}}]_{c=1..C}\eqno{(11)}$$
where $u \in \sigma={1,...,H}\times{1,...,W}$ is a 'pixel' in the map, $c$ is channel, $x_{cu}$ is the corresponding tensor element and $p$ is a learnable parameter. GeM pooling is a generation of the GAP($p=1$) and GMP($p=\infty$) and the larger the $p$ the more localized the feature map responses are. GeM pooling can learn the $p$ to aggregate the feature from the appropriate size area, which is between the 'whole image' and 'pixel', of the feature map. Therefore, we adopt GeM pooling in the HSA stream.

~\\\noindent\textbf{Performance comparison of global and head-shoulder features.} 
We demonstrate the capabilities of different features for person Re-ID in Table 6. We retrieve person images from Black-reID dataset using global feature, head-shoulder feature and integration of them in the experiment, which are denoted by HAA (ResNet50) (Global), HAA (ResNet50) (Head-shoulder) and HAA (ResNet50) (Global+Head-shoulder), respectively. From this table, we can observe two things: (1) The integration of global and head-shoulder feature achieves the best result in person Re-ID. HAA (ResNet50) (Global+Head-shoulder) gives $0.4\%$, $0.3\%$ and $34.2\%$, $35.2\%$ higher points in mAP and rank-1 accuracy than HAA (ResNet50) (Global) and HAA (ResNet50) (Head-shoulder), respectively. (2) Our model improves the representative capability of the global feature drastically. The improvements in mAP and rank-1 accuracy are $7.8\%$ and $5.5\%$ respectively.

\section{Conclusion}
We have presented a head-shoulder adaptive attention network to support person Re-ID with head-shoulder information. Through the adaptive attention module, the weights of the global and head-shoulder features can be automatically adjusted based on the type of the input person image. Our head-shoulder adaptive attention network can be integrated with the most current Re-ID models and is end-to-end trainable. We have also firstly proposed the Black Re-ID challenge and the first Black-reID dataset for further research. Our method not only achieves the best performance on Black-reID dataset but also on Market-1501 and DukeMTMC-reID and is proved to be effective in dealing with similar clothing. On Black-reID dataset, our model significantly outperforms the previous methods, by at least $\bm{+4.7\%/4.3\%}$ in mAP/rank-1 accuracy.

\section*{Acknowledgements}
This work was supported by the National Natural Science Foundation of China (Grant No. U1836217). The code was developed based on the 'fast-reid' toolbox \textcolor{blue}{https://github.com/JDAI-CV/fast-reid}.  

\bibliographystyle{ACM-Reference-Format}
\balance
\bibliography{ref}


\end{document}